\documentclass[a4paper]{llncs}

\usepackage{amsmath}
\usepackage{amsfonts}
\usepackage{epsfig}

\usepackage{isolatin1}
\usepackage[boxed]{algorithm}
\usepackage[noend]{algorithmic}

\newcommand{\reals}{\mathbb{R}}
\newcommand{\nats}{\mathbb{N}}
\newcommand{\reg}{\textup{\texttt{regular}}}
\newcommand{\softreg}{\textup{\texttt{soft\_regular}}}
\newcommand{\gcc}{\textup{\texttt{gcc}}}
\newcommand{\softgcc}{\textup{\texttt{soft\_gcc}}}
\newcommand{\softad}{\textup{\texttt{soft\_all\-dif\-fe\-rent}}}
\newcommand{\card}[1]{\left| #1 \right|}
\renewenvironment{proof}{\noindent{\bf Proof.}}{\mbox{}\hfill$\Box$\\}

\title{On Global Warming \\ (Softening Global Constraints)}
\author{Willem Jan van Hoeve\inst{1}, Gilles Pesant\inst{2,3}
  and Louis-Martin Rousseau\inst{2,3}}
\institute{CWI, P.O. Box 94079, 1090 GB Amsterdam, The Netherlands\\
\email{W.J.van.Hoeve@cwi.nl} 
\and
\'Ecole Polytechnique de Montr\'eal, Montreal, Canada 
\and
        Centre for Research on Transportation (CRT),\\ Universit\'e de
        Montr\'eal, C.P. 6128, succ. Centre-ville, Montreal, H3C 3J7,
        Canada \\
\email{ \{pesant,louism\}@crt.umontreal.ca}
}

\begin{document}
\maketitle

\begin{abstract}
We describe soft versions of the global cardinality constraint and 
the regular constraint, with efficient filtering algorithms maintaining 
domain consistency. For both constraints, the softening is achieved by 
augmenting the underlying graph. The softened constraints can be used to
extend the meta-constraint framework for over-constrained problems 
proposed by Petit, R\'egin and Bessi\`ere.\end{abstract}

\section{Introduction}
\label{intro}

Constraint Programming (CP) is a widely used and efficient technique to
solve combinatorial optimization problems. However in practice many
problems are over-constrained (intrinsically or from being badly stated). 
Several frameworks have been proposed to handle over-constrained problems,
mostly by introducing {\em soft constraints} that are allowed to be 
(partially) violated. 
The most well-known framework is the Partial Constraint 
Satisfaction Problem framework (PCSP \cite{PCSP}), which includes the 
Max-CSP framework that tries to maximize the number of satisfied 
constraints. Since in this framework all constraints are either violated 
or satisfied, this objective is equivalent to minimizing the number of
violations. It has been extended to the \emph{Weighted-CSP}
\cite{larrosa2002,larrosa2003}, associating a degree of violation
 (not just a boolean value) to each constraint and minimizing the
sum of all weighted violations. The \emph{Possibilistic-CSP}
\cite{PossibilisticCSP} associates a preference to each constraint (a
real value between 0 and 1) representing its importance. The objective
of the framework is the hierarchical satisfaction of the most
important constraints, that is, the minimization of the highest preference level for a violated constraint. 
The \emph{Fuzzy-CSP}
\cite{FuzzyCSP,fargier1993} is somewhat similar to the
Possibilistic-CSP but here a preference is associated to each tuple of
each constraint. A preference value of 0 means the constraint is
highly violated and 1 stands for satisfaction. The objective is the
maximization of the smallest preference value induced by a variable assignment.
The last two frameworks
are different from the previous ones since the aggregation operator is
a $min/max$ function instead of addition. Max-CSPs are typically
encoded and solved with one of two generic paradigms: valued-CSPs 
\cite{schiex1995} and semi-rings \cite{bistarelli}. 

Another approach to model and solve over-constrained problems involves
\emph{Meta-Constraints} \cite{MetaConstraints}. The idea behind
this technique is to introduce a set of domain variables $Z$ that capture the
violation cost of each soft constraint. By correctly constraining
these variables it is possible to replicate the previous frameworks and
even to extend the modeling capability to capture other types of
violation measures. Namely the authors argue that although the Max-CSP
family of frameworks is quite efficient to capture local violation
measures it is not as adequate to model violation costs involving
several soft constraints simultaneously. By defining (possibly global)
constraints on $Z$ such a behaviour can be easily achieved. The authors
propose to replace each soft constraint $S_i$ present in a model by a
disjunctive constraint specifying that either $z_i=0$ and the constraint
$S_i$ is hard or $z_i>0$ and $S_i$ is violated. This technique allows
the resolution of over-constrained problem within traditional CP
solvers. 

Comparatively few efforts have been invested in developing soft
versions of common global constraints
\cite{petit:cp01,beldi-petit:cpaior04,vanhoeve2004}.
Global constraints are often
key elements in successfully modeling real applications and being able to
easily and effectively soften such constraints would yield a significant 
improvement in flexibility.
In this paper we study two global constraints: the widely known global
cardinality constraint (\gcc) \cite{regin96} and the new \reg\
\cite{regular04} constraint. For each of these we propose new
violation measures and provide the corresponding filtering algorithms
to achieve domain consistency. All the constraint softening is
achieved by enriching the underlying graph representation with
additional arcs that represent possible relaxations of the
constraint. Violation costs are then associated to these new arcs and
known graph algorithms are used to achieve domain consistency. 
      
The two constraints studied in this paper are useful to model and 
solve personnel rostering problems (PRP). The PRP objective is typically to
distribute a set of working shifts (or days off) to a set of employees every
day over a planning horizon (a set of days). The \gcc\ is a perfect tool to 
restrict the number
of work shifts of each type (Day, Evening, and Night for instance) performed
by each employee. Other types of constraints involve sequences of shifts over 
time, typically forbidding non ergonomic schedules.
The \reg\ constraint has the expressive power necessary to cope with
the complex regulations found in many organizations. Since most real
rostering applications are over-constrained (due to lack of personnel or
over-optimistic scheduling objectives), soft versions of the \gcc\ and \reg\
constraints promise to significantly improve our modelling flexibility.

This paper is organized as follows. Section~\ref{background} presents 
background information on Constraint Programming and the softening of 
(global) constraints. In Section~\ref{gcc} and \ref{reg} we describe the 
softening of the \gcc\ and the \reg\ constraint respectively. Both 
constraints are softened with respect to two violation measures. We also 
provide corresponding filtering algorithms achieving domain consistency.
Section~\ref{agg} discusses the aggregation of several soft (global) 
constraints by meta-constraints. Finally, a conclusion is given in
Section~\ref{conclusion}.

\section{Background}
\label{background}

We assume familiarity with the basic concepts of constraint
programming. For a thorough explanation of constraint programming,
see~\cite{apt2003}.

A constraint satisfaction problem (CSP) consists of a finite set of
variables $X = \{ x_1, \dots , x_n \}$ with finite domains 
$\mathcal{D} = \{ D_1, \dots, D_n \}$ such that $x_i \in D_i$ for all 
$i$, together with a finite set of constraints $\mathcal{C}$, each on 
a subset of $X$. 
A constraint $C \in \mathcal{C}$ is defined as a subset of the 
Cartesian product of the domains of the variables that are in $C$. 
A tuple $(d_1, \dots, d_n) \in D_1 \times \dots \times D_n$ is a
solution to a CSP if for every constraint $C \in \mathcal{C}$ on the 
variables $x_{i_1}, \dots , x_{i_k}$ we have 
$(d_{i_1}, \dots, d_{i_k}) \in C$. A constraint optimization problem
(COP) is a CSP together with an objective function to be optimized. A
solution to a COP is a solution to the corresponding CSP that has an
optimal objective function value.

\begin{definition}[Domain consistency]\label{def:hac}
A constraint $C$ on the variables $x_1, \dots,$ $x_k$ is called 
domain consistent if for each variable $x_i$ and value
$d_i \in D_i$, there exist values $d_1,\dots, d_{i-1}, d_{i+1}, \dots, d_k$ 
in $D_1, \dots, D_{i-1},D_{i+1}, \dots, D_k$, such that $(d_1, \dots , d_k) 
\in C$. 
\end{definition}
Our definition of domain consistency corresponds to hyper-arc consistency 
or generalized arc consistency, which are also often used in the literature.

\begin{definition}[Consistent CSP]
A CSP is domain consistent if all its constraints are domain
consistent. A CSP is inconsistent if it has no solution. Similarly for
a COP.
\end{definition}

When a CSP is inconsistent it is also said to be over-constrained. It is 
then natural to identify soft constraints, that are allowed to be violated,
and minimize the total violation according to some criteria. For each
soft constraint $C$, we introduce a function that measures the violation, 
and has the following form:
\begin{displaymath}
{\rm violation}_C: D_1 \times \cdots \times D_n \rightarrow \nats.
\end{displaymath} 
This approach has been introduced in \cite{petit:cp01} and was developed 
further in \cite{beldi-petit:cpaior04}. There may be several natural ways 
to evaluate the degree to which a global constraint is violated and these
are not equivalent usually. A standard measure is the variable-based cost:
\begin{definition}[Variable-based cost]\label{def:varcost}
Given a constraint $C$ on the variables $x_1, \dots, x_k$ and an
instantiation $d_1, \dots, d_k$ with $d_i \in D_i$, the variable-based 
cost of violation of $C$ is the minimum number of variables that need to
change their value in order to satisfy the constraint.
\end{definition}
Alternative measures exist for specific constraints. For example, if a 
constraint is expressible as a conjunction of binary constraints, the 
cost may be defined as the number of these binary constraints that are 
violated. For the soft \gcc\ and the soft \reg\ constraint, we will
introduce new violation measures, that are likely to be more effective 
in practical applications.

\section{Soft Global Cardinality Constraint}
\label{gcc}
A global cardinality constraint (\gcc) on a set of variables
specifies the minimum and maximum number of times each value in the
union of their domains should be assigned to these variables.
R\'egin developed a domain consistency algorithm for the \gcc,
making use of network flows \cite{regin96}. A variant of the \gcc\ is
the cost-\gcc, which can be seen as a weighted version of the \gcc\
\cite{regin99,regin2002}. For the cost-\gcc\ a weight is assigned to 
each variable-value assignment and the goal is to satisfy the \gcc\ 
with minimum total cost.

Throughout this section, we will use the following notation (unless
specified otherwise). Let $X$ denote a set of variables 
$\{x_1, \dots, x_n\}$ with respective finite domains $D_1, \dots, D_n$. 
We define $D_X = \cup_{i \in \{1, \dots, n\}} D_i$ and we assume a fixed 
but arbitrary ordering on $D_X$. For $d \in D_X$, let $l_d, u_d \in \nats$, 
with $l_d \leq u_d$. Finally, let $z$ be a variable with finite domain $D_z$,
representing the cost of violation of the \gcc.

\begin{definition}[Global cardinality constraint]\label{def:gcc}
\begin{displaymath}
\gcc(X, l, u) = 
\{(d_1, \dots, d_n) \mid d_i \in D_i, l_d \leq \card{ \{ d_i \mid d_i = d \} }
\leq u_d \; \forall \; d \in D_X \}.
\end{displaymath}
\end{definition}
We first give a generic definition for a soft version of the \gcc.
\begin{definition}[Soft global cardinality constraint]
\label{def:softgcc}
\begin{displaymath}
\begin{array}{ll}
\softgcc[\star](X, l, u, z) = 
\{(d_1, \dots, d_n, \tilde{d}) \mid 
& d_i \in D_i, \tilde{d} \in D_z, \\
& {\rm violation}_{\softgcc[\star]}(d_1, \dots, d_n) \leq \tilde{d} \},
\end{array}
\end{displaymath}
where $\star$ defines a violation measure for the \gcc.
\end{definition}
In order to define measures of violation for the \gcc, it is
convenient to introduce the following functions.
\begin{definition}[Overflow, underflow]
Given \gcc$(X, l, u)$, define for all $d \in D_X$
\begin{displaymath}
{\rm overflow}(X, d) = \left\{
\begin{array}{ccl}
\card{ \{x_i \mid x_i = d \} } - u_d & \hspace{2em} & {\rm if} \; 
    \card{ \{ x_i \mid x_i = d \} } \geq u_d,\\
0 & & {\rm otherwise},
\end{array}\right.
\end{displaymath}
\begin{displaymath}
{\rm underflow}(X, d) = \left\{
\begin{array}{ccl}
l_d - \card{ \{ x_i \mid x_i = d \} } & \hspace{2em} & {\rm if} \; 
    \card{ \{ x_i \mid x_i = d \} } \leq l_d,\\
0 & & {\rm otherwise}.
\end{array}\right.
\end{displaymath}
\end{definition}
Let ${\rm violation}_{\softgcc[{\rm var}]}$ denote the variable-based 
cost of violation (see Definition~\ref{def:varcost}) of the \gcc.
The next lemma expresses ${\rm violation}_{\softgcc[{\rm var}]}$ in 
terms of the above functions.
\begin{lemma}\label{lem:varcost}
Given \gcc$(X, l, u)$,
\begin{displaymath}
{\rm violation}_{\softgcc[{\rm var}]}(X) = 
\max \left( \sum_{d \in D_X} {\rm overflow}(X, d) , \sum_{d \in D_X}
{\rm underflow}(X, d) \right)
\end{displaymath}
provided that 
\begin{equation}\label{eq:assumption}
\sum_{d \in D_X} l_d \leq \card{X} \leq \sum_{d \in D_X}
u_d. 
\end{equation}
\end{lemma}
\begin{proof}
The variable-based cost of violation corresponds to the minimal 
number of re-assignments of variables until both 
$\sum_{d \in D_X} {\rm overflow}(X, d) = 0$ and $\sum_{d \in D_X} 
{\rm underflow}(X, d) = 0$.

Assume $\sum_{d \in D_X} {\rm overflow}(X, d) \geq \sum_{d \in D_X} 
{\rm underflow}(X, d)$. Variables assigned to values 
$d^\prime \in D_X$ with ${\rm overflow}(X,d^\prime) > 0$ can be 
assigned to values $d^{\prime\prime} \in D_X$ with 
${\rm underflow}(X,d^{\prime\prime}) > 0$, until 
$\sum_{d \in D_X} {\rm underflow}(X,d) = 0$. In order to achieve 
$\sum_{d \in D_X} {\rm overflow}(X, d) = 0$, we still need to 
re-assign the other variables assigned to values 
$d^\prime \in D_X$ with ${\rm overflow}(X,d^\prime) > 0$. 
Hence, in total we need to re-assign exactly 
$\sum_{d \in D_X} {\rm overflow}(X, d)$ variables.

Similarly when we assume $\sum_{d \in D_X} {\rm overflow}(X, d) \leq 
\sum_{d \in D_X} {\rm underflow}(X, d)$.

If (\ref{eq:assumption}) does not hold, there is no variable
assignment that satisfies the \gcc.
\end{proof}\\
Without assumption (\ref{eq:assumption}), the variable-based
violation measure for the \gcc\ cannot be applied. Therefore, we 
introduce the following value-based violation measure, which can 
also be applied when assumption (\ref{eq:assumption}) does not hold.
\begin{definition}[Value-based cost]\label{def:valcost}
For \gcc$(X, l, u)$ the value-based cost of violation is
\begin{displaymath}
\sum_{d \in D_X} {\rm overflow}(X,d) + {\rm underflow}(X,d).
\end{displaymath}
\end{definition}
We denote the value-based violation measure for the \gcc\ by 
${\rm violation}_{\softgcc[{\rm val}]}$.

\subsection{Graph Representation}
First, we introduce the concept of a flow in a directed graph, 
following Schrij\-ver~\cite[pp. 148--150]{lex2003}.

A directed graph is a pair ${\cal G} = (V,A)$ where $V$ is a finite set of 
vertices and $A$ is a family\footnote{A family is a set in which 
elements may occur more than once.} of ordered pairs from $V$, called 
arcs. For $v \in V$, let $\delta^{\rm in}(v)$ and $\delta^{\rm out}(v)$ 
denote the family of arcs entering and leaving $v$ respectively.

A (directed) walk in $\cal G$ is a sequence $P = v_0, a_1, v_1, \dots, a_k,
v_k$ where $k \geq 0$, $v_0, v_1, \dots, v_k \in V$, $a_1, \dots, a_k
\in A$ and $a_i = (v_{i-1}, v_i)$ for $i = 1, \dots, k$. If there is
no confusion, $P$ may be denoted as $P = v_0, v_1, \dots, v_k$. A
(directed) walk is called a (directed) path if  $v_0, \dots, v_k$ are
distinct. A closed (directed) walk, i.e. $v_0 = v_k$, is called a
(directed) circuit if $v_1, \dots, v_k$ are distinct.

Let $s,t \in V$. We apply a capacity function 
$c:A \rightarrow \reals_+$, a demand function $d:A \rightarrow \reals_+$ 
and a cost function $w:A \rightarrow \reals_+$ on the arcs.
A function $f: A \rightarrow \reals$ is called a \emph{feasible flow} from 
$s$ to $t$, or an $s-t$ flow, if
\begin{eqnarray}
d(a) \leq f(a) \leq c(a) & \textrm{for each } a \in A, \\ 
f(\delta^{\rm out}(v)) = f(\delta^{\rm in}(v)) & 
\textrm{for each } v \in V \setminus \{s,t\}, \label{eq:flow}
\end{eqnarray}
where $f(S) = \sum_{a \in S} f(a)$ for all $S \subseteq A$.
Property (\ref{eq:flow}) ensures flow conservation, i.e. for
a vertex $v \neq s,t$, the amount of flow entering $v$ is equal to the 
amount of flow leaving $v$. 
The value of an $s-t$ flow $f$ is defined as
\begin{displaymath}
{\rm value}(f) = f(\delta^{\rm out}(s)) - f(\delta^{\rm in}(s)).
\end{displaymath}
In other words, the value of a flow is the net amount of flow leaving $s$, 
which can be shown to be equal to the net amount of flow entering $t$.
The cost of a flow $f$ is defined as 
\begin{displaymath}
{\rm cost}(f) = \sum_{a \in A} w(a) f(a).
\end{displaymath}
A \emph{minimum-cost flow} is a feasible $s-t$ flow of minimum cost. The 
\emph{minimum-cost flow problem} is the problem of finding such a minimum-cost 
flow.
\begin{theorem}[\cite{regin96}]
A solution to \gcc$(X, l, u)$ corresponds to a feasible $s-t$ flow of 
value $n$ in the graph ${\cal G} = (V, A)$ with vertex set 
\begin{displaymath}
V = X \cup D_X \cup \{s,t\}
\end{displaymath}
and edge set
\begin{displaymath}
A = A_{s \rightarrow X} \cup A_{X \rightarrow D_X} \cup A_{D_X
  \rightarrow t},
\end{displaymath}
where
\begin{displaymath}
\begin{array}{rl}
A_{s \rightarrow X} =& \{ (s, x_i) \mid i \in \{1, \dots, n \} \}, \\ 
A_{X \rightarrow D_X} =& \{ (x_i, d) \mid d \in D_i, i \in \{1, \dots,
n \} \}, \\
A_{D_X \rightarrow t} =& \{ (d, t) \mid d \in D_X \},
\end{array}
\end{displaymath}
with demand function
\begin{displaymath}
d(a) = \left\{
\begin{array}{ll}
1 & {\rm if} \; a \in A_{s \rightarrow X}, \\ 
0 & {\rm if} \; a \in A_{X \rightarrow D_X}, \\
l_d & {\rm if} \; a = (d,t) \in A_{D_X \rightarrow t},
\end{array} \right.
\end{displaymath}
and capacity function
\begin{displaymath}
c(a) = \left\{
\begin{array}{ll}
1 & {\rm if} \; a \in A_{s \rightarrow X}, \\ 
1 & {\rm if} \; a \in A_{X \rightarrow D_X}, \\
u_d & {\rm if} \; a = (d,t) \in A_{D_X \rightarrow t}.
\end{array} \right.
\end{displaymath}
\end{theorem}

\begin{figure}[t]
\begin{center}
\begin{minipage}[t]{.3\textwidth}
\begin{center}
\epsfig{figure=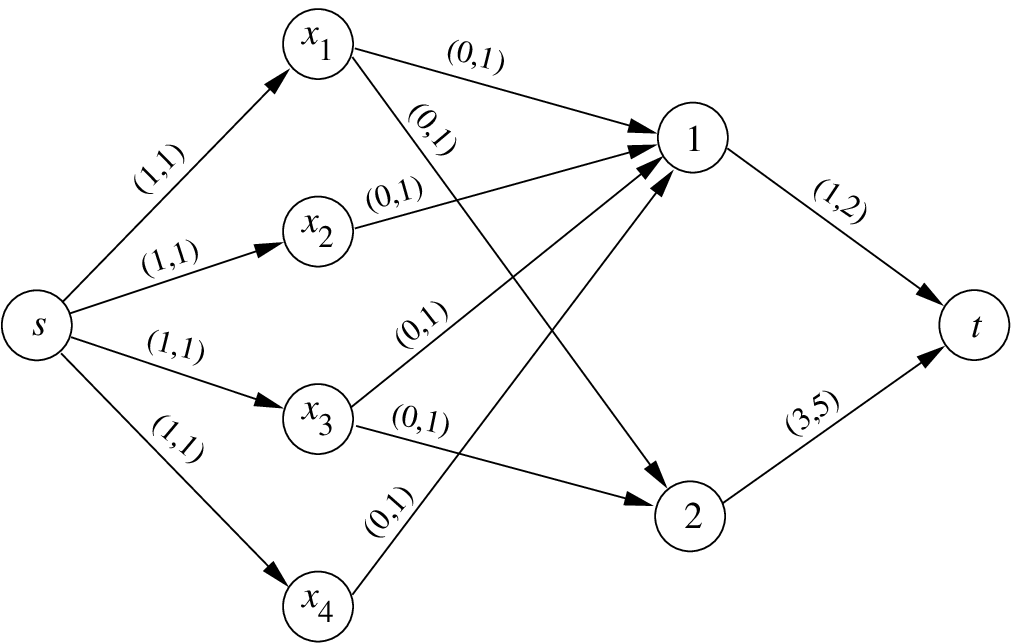,width=\textwidth} \\
a.~~original \gcc
\end{center}
\end{minipage} \hspace{1ex}
\begin{minipage}[t]{.3\textwidth}
\begin{center}
\epsfig{figure=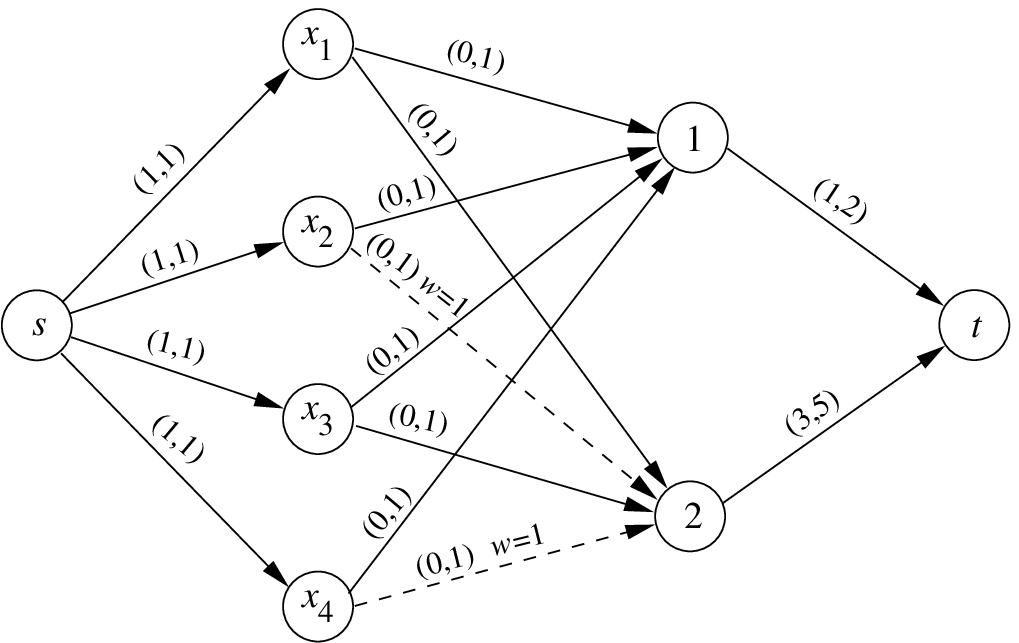,width=\textwidth} \\
b.~~\softgcc[{\rm var}]
\end{center}
\end{minipage} \hspace{1ex}
\begin{minipage}[t]{.3\textwidth}
\begin{center}
\epsfig{figure=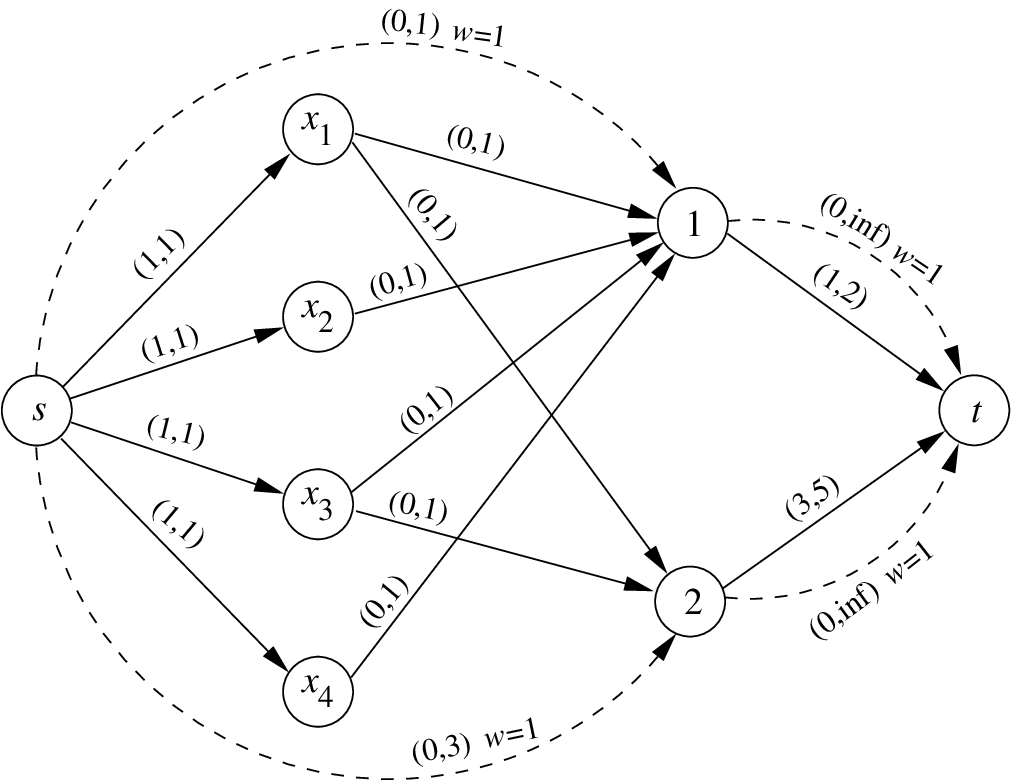,width=\textwidth} \\
c.~~\softgcc[{\rm val}]
\end{center}
\end{minipage}
\end{center}
\caption{Graph representation for the \gcc, the variable-based \softgcc\
  and the value-based \softgcc. Demand and capacity are indicated between 
  parentheses for each arc. Dashed arcs indicate the inserted weighted 
  arcs.}
\label{fig:gcc}
\end{figure}

\begin{example}
Consider the CSP
\begin{displaymath}
\begin{array}{l}
x_1 \in \{1, 2\}, x_2 \in \{1\},  x_3 \in \{1, 2\}, x_4 \in \{1\},\\
\gcc(X, l, u)
\end{array}
\end{displaymath}
where $X = \{x_1, \dots, x_4\}$, $l_1 = 1$, $l_2 = 3$, $u_1 = 2$ and
$u_2 = 5$. In Figure~\ref{fig:gcc}.a the corresponding graph ${\cal G}$ for the
\gcc\ by applying the above procedure is presented. 
\end{example}


\subsection{Variable-Based Violation}
For the variable-based violation measure, we adapt the graph ${\cal G}$ in the
following way. We add the arc set 
$\tilde{A}_{X \rightarrow D_X} = 
  \{ (x_i, d) \mid d \notin D_i, i \in \{1, \dots, n \} \}$,
with demand $d(a) = 0$, capacity $c(a) = 1$ for all arcs 
$a \in \tilde{A}_{X \rightarrow D_X}$.
Further, we apply a cost function $w: A \rightarrow \reals$, where
\begin{displaymath}
w(a) = \left\{
\begin{array}{ll}
1 & {\rm if} \; a \in \tilde{A}_{X \rightarrow D_X},\\
0 & {\rm otherwise}.
\end{array} \right.
\end{displaymath}
Let the resulting graph be denoted by ${\cal G}_{\rm var}$.
\begin{example}
Consider the CSP
\begin{displaymath}
\begin{array}{l}
x_1 \in \{1, 2\}, x_2 \in \{1\},  x_3 \in \{1, 2\}, x_4 \in \{1\}, 
z \in \{0, 1, \dots, 4\} \\
\softgcc[{\rm var}](X, l, u, z)\\
\texttt{minimize } z
\end{array}
\end{displaymath}
where $X = \{x_1, \dots, x_4\}$, $l_1 = 1$, $l_2 = 3$, $u_1 = 2$ and
$u_2 = 5$. In Figure~\ref{fig:gcc}.b the graph ${\cal G}_{\rm var}$ for the 
\softgcc[{\rm var}] is presented.
\end{example}

\begin{theorem}\label{thm:var}
A minimum-cost flow in the graph ${\cal G}_{\rm var}$ corresponds to
a solution to the $\softgcc[{\rm var}]$, minimizing the variable-based 
violation.
\end{theorem}
\begin{proof}
An assignment $x_i = d$ corresponds to the arc $a = (x_i, d)$ with 
$f(a) = 1$. By construction, all variables need to be assigned to 
a value and the cost function exactly measures the variable-based cost 
of violation.
\end{proof}

The graph ${\cal G}_{\rm var}$ corresponds to a particular instance of the
cost-\gcc\ \cite{regin99,regin2002}. Hence, we can apply the filtering
procedures developed for that constraint directly to the \softgcc[{\rm var}].
The \softgcc[{\rm var}] also inherits from the cost-\gcc\ the time complexity 
of achieving domain consistency, being $O(n(m + n \log n))$ where 
$m = \sum_{i=1}^n \card{D_i}$ and $n = \card{X}$.

Note that \cite{beldi-petit:cpaior04} also consider the variable-based 
cost measure for a different version of the soft \gcc. Their version
considers the parameters $l$ and $u$ to be variables too. Hence, the 
variable-based cost evaluation becomes a rather poor measure, as we 
trivially can change $l$ and $u$ to satisfy the \gcc. They fix this by 
restricting the set of variables to consider to be the set $X$,
which corresponds to our situation. However, they do not provide a 
filtering algorithm for that case.

\subsection{Value-Based Violation}
For the value-based violation measure, we
adapt the graph ${\cal G}$ in the following way. We add 
arc sets $A_{\rm underflow} = \{ (s, d) \mid d \in D_X \}$
and $A_{\rm overflow} = \{ (d, t) \mid d \in D_X \}$,
with demand $d(a) = 0$ for all $a \in A_{\rm underflow} \cup 
A_{\rm overflow}$ and capacity
\begin{displaymath}
c(a) = \left\{
\begin{array}{ll}
l_d & {\rm if} \; a = (s,d) \in A_{\rm underflow},\\
\infty & {\rm if} \; a \in A_{\rm overflow}.\\
\end{array} \right.
\end{displaymath}
Further, we again apply a cost function $w: A \rightarrow \reals$,
where 
\begin{displaymath}
w(a) = \left\{
\begin{array}{ll}
1 & {\rm if} \; a \in A_{\rm underflow} \cup A_{\rm overflow},\\
0 & {\rm otherwise}.
\end{array} \right.
\end{displaymath}
Let the resulting graph be denoted by ${\cal G}_{\rm val}$.
\begin{example}
Consider the CSP
\begin{displaymath}
\begin{array}{l}
x_1 \in \{1, 2\}, x_2 \in \{1\},  x_3 \in \{1, 2\}, x_4 \in \{1\}, 
z \in \{0, 1, \dots, 5\} \\
\softgcc[{\rm val}](X, l, u, z)\\
\texttt{minimize } z
\end{array}
\end{displaymath}
where $X = \{x_1, \dots, x_4\}$, $l_1 = 1$, $l_2 = 2$, $u_1 = 3$ and
$u_2 = 2$. In Figure~\ref{fig:gcc}.c the graph ${\cal G}_{\rm val}$ for the 
\softgcc\ with respect to value-based cost is presented.
\end{example}

\begin{theorem}
A minimum-cost flow in the graph ${\cal G}_{\rm val}$ corresponds to
a solution to the $\softgcc[{\rm val}]$, minimizing the value-based 
violation.
\end{theorem}
\begin{proof}
An assignment $x_i = d$ corresponds to the arc $a = (x_i, d)$ 
with $f(a) = 1$. By construction, all variables need to be assigned to 
a value and the cost function exactly measures the value-based cost of
violation.
\end{proof}

Unfortunately, the graph ${\cal G}_{\rm val}$ does not preserve the 
structure of the cost-\gcc\ because of the arcs $A_{\rm underflow}$. 
Therefore we cannot blindly apply the same filtering algorithms.
However, it is still possible to design an efficient filtering algorithm 
for the value-based \softgcc\ (in the same spirit of the filtering 
algorithm for the cost-\gcc), based again on flow theory.
For this, we need to introduce the residual graph
${\cal G}^f = (V,A^f)$ of a flow $f$ on ${\cal G} = (V,A)$ (with respect 
to $c$ and $d$), where 
\begin{displaymath}
A^f = \{a \mid a \in A, f(a) < c(a)\} \cup 
\{a^{-1} \mid a \in A, f(a) > d(a) \}.
\end{displaymath}
Here $a^{-1} = (v,u)$ if $a = (u,v)$. We extend $w$ to 
$A^{-1} = \{a^{-1} \mid a \in A\}$ by defining
$w(a^{-1}) = - w(a)$
for each $a \in A$.
\begin{theorem}
Let $f$ be a minimum-cost flow in ${\cal G}_{\rm val}$. 
Then $\softgcc[{\rm val}](X,l,u,z)$ is domain consistent if and 
only if \[\min{D_z} \geq {\rm cost}(f)\] and 
\begin{displaymath}
\begin{array}{ll}
{\rm cost}(f) + {\rm cost}({\rm SP}(d, x_i)) \leq \max{D_z} &
\forall (x_i,d) \in A_{X \rightarrow D_X},
\end{array}
\end{displaymath}
where ${\rm cost}({\rm SP}(d, x_i))$ denotes the cost of a shortest path
from $d$ to $x_i$ in the residual graph ${\cal G}^f_{\rm val}$.
\end{theorem}
\begin{proof}
From flow theory \cite{ahuja} we know that, given a minimum-cost flow $f$
in ${\cal G}_{\rm val}$, if we enforce arc $(x_i, d)$ to be in a minimum-cost 
flow $\tilde{f}$ in ${\cal G}_{\rm val}$, ${\rm cost}(\tilde{f}) = 
{\rm cost}(f) + {\rm cost}({\rm SP}(d, x_i))$ where ${\rm SP}(d, x_i)$ is the 
shortest $d-x_i$ path in ${\cal G}_{\rm val}^f$.

In order for a value $d \in D_i$ to be consistent, the cost of a
minimum-cost flow that uses $(x_i, d)$ should be less than or equal to
$\max{D_z}$. By the above fact, we only need to compute a shortest
path from $d$ to $x_i$ instead of a new minimum-cost flow.
\end{proof}\\
A minimum-cost flow $f$ in ${\cal G}_{\rm val}$ can be computed in 
$O(m (m + n \log n))$ time (see \cite{ahuja}), where again 
$m = \sum_{i=1}^n \card{D_i}$ and $n = \card{X}$. 
Compared to the complexity of 
the \softgcc[{\rm var}], we have a factor $m$ instead of $n$. This is because
computing the flow for \softgcc[{\rm val}] is dependent on the number of arcs 
$m$ rather than on the number variables $n$.
A shortest $d-x_i$ path in ${\cal G}_{\rm val}$ can be computed in 
$O(m + n \log n)$ time. Hence the \softgcc\ with respect 
to the value-based violation measure can be made domain 
consistent in $O((m-n) (m + n \log n))$ time as we need to check 
$m-n$ arcs for consistency.

When $l = \vec{0}$ in $\softgcc[{\rm val}](X, l, u, z)$, the arc set
$A_{\rm underflow}$ is empty. In that case, ${\cal G}_{\rm val}$ has a 
particular structure, i.e. the only costs appear on arcs from $D_X$ to 
$t$. As pointed out in \cite{vanhoeve2004} for the \softad\ constraint, 
constraints with this structure can be checked for consistency in 
$O(nm)$ time, and domain consistency can be achieved in $O(m)$ time.
The result is obtained by exploiting the strongly connected 
components\footnote{A strongly connected component in a directed graph 
${\cal G} = (V, A)$ is a subset of vertices $S \subseteq V$ such that 
there exists a directed $u-v$ path in ${\cal G}$ for all $u, v \in S$.}
in ${\cal G}_{\rm val}$ restricted to vertex sets $X$ and $D_X$.

\section{Soft Regular Constraint}
\label{reg}

A \reg\ constraint \cite{regular04} on a fixed-length sequence of
finite-domain variables
requires that the corresponding sequence of values taken by these variables
belong to a given regular language.
A \emph{deterministic finite automaton} (DFA) may be described by a $5$-tuple
$M=(Q,\Sigma,\delta,q_0,F)$ where
$Q$ is a finite set of states, 
$\Sigma$ is an alphabet, 
$\delta: Q \times \Sigma \rightarrow Q$ is a partial transition function,
$q_0 \in Q$ is the initial state, and
$F \subseteq Q$ is the set of final (or accepting) states.
A finite sequence of symbols from an alphabet is called a
\emph{string}.
Strings processed by $M$ and ending in an accepting state from $F$ are said
to belong to the language defined by $M$, denoted $L(M)$.
The languages recognized by DFAs are precisely regular languages.

Given a sequence ${\mathbf x}=\langle  x_1,x_2,\ldots,x_{n} \rangle$
of finite-domain variables with respective domains $D_1$, $D_2$,
\ldots, $D_n \subseteq \Sigma$, there is a natural interpretation of
the set of possible instantiations of ${\mathbf x}$, $D_1 \times D_2
\times \cdots \times D_n$, as a subset of all strings of length $n$
over $\Sigma$, $\Sigma^n$.
We are now ready to state the constraint.

\begin{definition}[Regular language membership constraint]
Let ${M} = $ \linebreak $(Q,\Sigma,\delta,q_0,F)$ denote a deterministic
finite automaton and $\mathbf x$ a sequence of finite-domain variables
$\langle  x_1,x_2,\ldots,x_{n} \rangle$ with respective domains $D_1$, $D_2$,
\ldots, $D_n \subseteq \Sigma$.
Under a \emph{regular language membership constraint} \linebreak
{\tt regular(}${\mathbf x},M${\tt )}, 
any sequence of values taken by the variables of ${\mathbf x}$
corresponds to a string in $L(M)$.
\end{definition}

In \cite{regular04}, a domain consistency algorithm for the \reg\
constraint processed the sequence ${\mathbf x}$ with the automaton
$M$, 
building a layered
directed multi-graph ${\cal G}=(N^1,N^2,\ldots,N^{n+1},A)$ where each layer $N^i =
\{q^i_0, q^i_1, \ldots, q^i_{|Q|-1}\}$ contains a different node for
each state of $M$ and arcs only appear between consecutive layers.
Each arc corresponds to a consistent variable-value pair:
there is an arc from $q^i_k$ to $q^{i+1}_{\ell}$ if and only
if there exists some $v_j \in D_i$ such that $\delta(q_k,v_j) =
q_{\ell}$ and the arc belongs to a path from 
$q_0$ in the first layer to a member of $F$ in the last layer.
The existence of such an arc, labeled $v_j$, constitutes a support for
variable $x_i$ taking value $v_j$.

\begin{figure}
\begin{center}
~
\psfig{figure=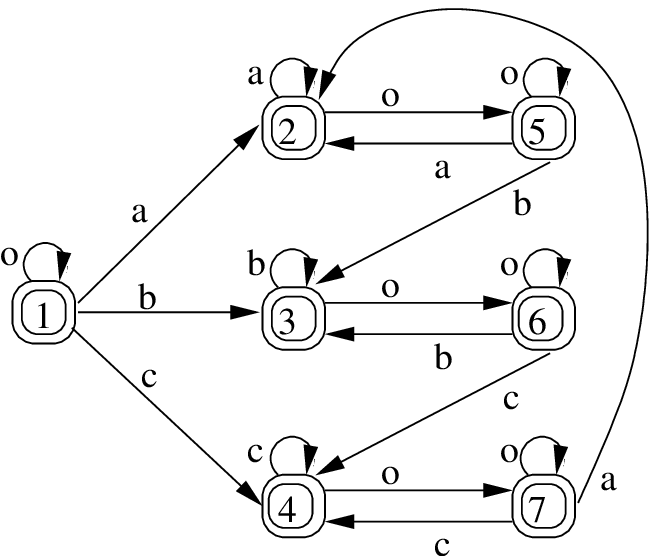,height=3.7cm}
\hspace{2em}
\psfig{figure=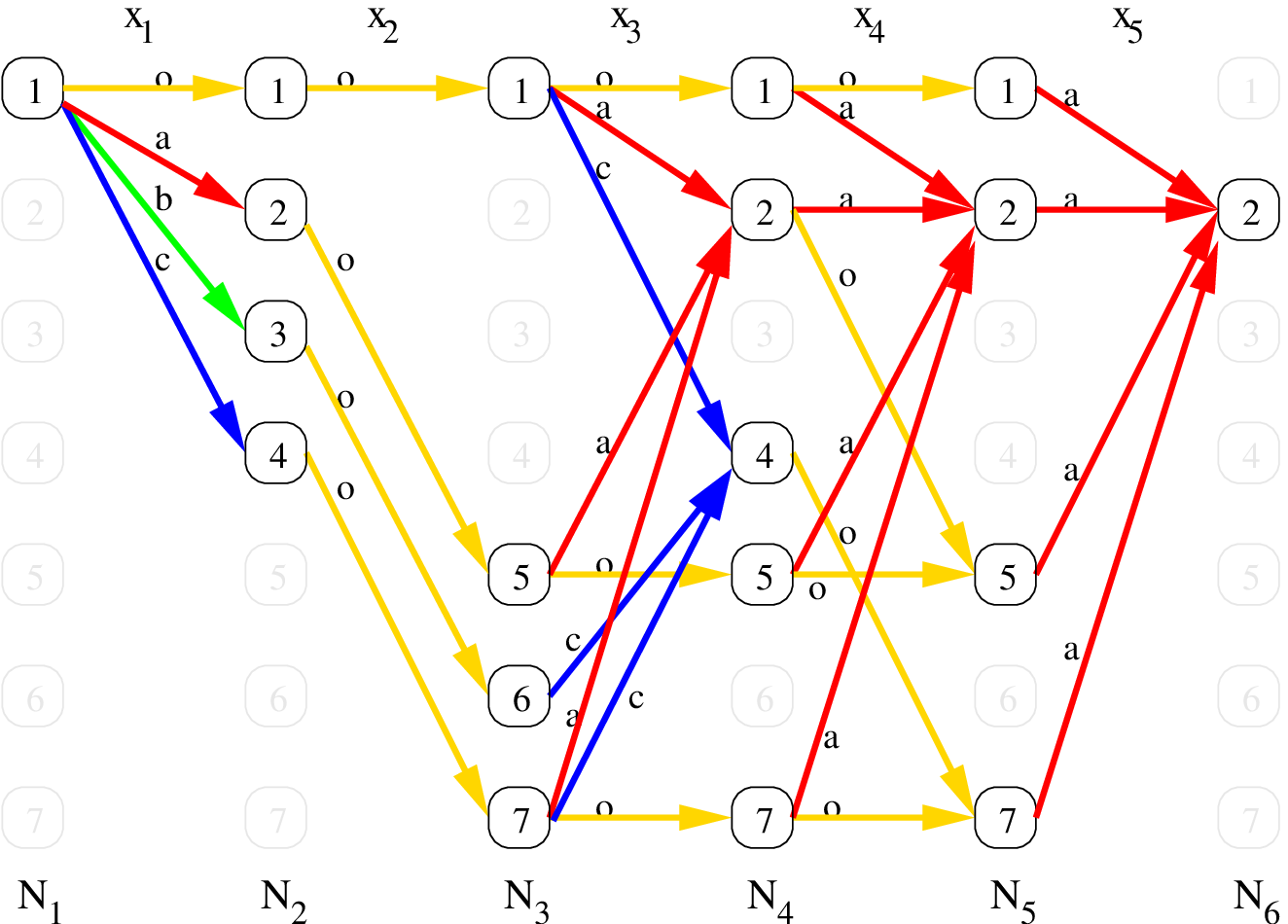,width=5.5cm}
\end{center}
\caption{A DFA (left) and its layered directed graph ${\cal G}$ (right).}
\label{digraph}
\end{figure}

For example, consider a sequence ${\mathbf x}$ of five variables with 
$D_1 = \{a,b,c,o\}$, $D_2 = \{b,o\}$, $D_3 = \{a,c,o\}$, $D_4 =
\{a,b,o\}$, and $D_5 = \{a\}$.
Figure \ref{digraph} gives an automaton $M$ (with its initial state labeled $1$) and the resulting graph for
constraint {\tt regular(}${\mathbf x},M${\tt )}.
As a result, value $b$ is removed from $D_2$ and $D_4$.

\subsection{Cost Definition}

We first give a generic definition for a soft version of the \reg\ constraint. 

\begin{definition}[Soft regular language membership constraint] 
Let ${M} = $ $(Q,\Sigma,\delta,q_0,F)$ denote a deterministic
finite automaton and $\mathbf x$ a sequence of finite-domain variables
$\langle  x_1,x_2,\ldots,x_{n} \rangle$ with respective domains $D_1$, $D_2$,
\ldots, $D_n \subseteq \Sigma$.
Let $z$ be a finite-domain variable of domain $D_z \subset \nats$
representing the cost of a
violation and let $d:\Sigma^{\star} \times \Sigma^{\star} \rightarrow
\nats$ be some distance function over strings.
Under a \emph{soft regular language membership constraint} 
$\softreg[d]({\mathbf x},M,z)$, 
for any sequence of values $\sigma$ taken by the variables of
${\mathbf x}$ we have \linebreak
$\min_{\sigma' \in L(M)} \{d(\sigma,\sigma')\} = z$.
\end{definition}

Our first instantiation of the distance function yields the
variable-based cost:
\begin{definition}[Hamming distance]
The number of positions in which two strings of same length differ is
called their \emph{Hamming distance}.  
\end{definition}
Intuitively, such a distance represents the number of symbols we need
to change to go from one string to the other, or equivalently the
number of variables whose value must change.
Using the Hamming distance for $d$ in the previous definition, $z$
becomes the variable-based cost.
 
Another distance function that is often used with strings is the following:
\begin{definition}[Edit distance]
The smallest number of insertions, deletions, and substitutions
required to change one string into another is called the
\emph{edit distance}.  
\end{definition}
It captures the fact that two strings that are identical except for
one extra or missing symbol should be considered close to one another. 
For example, the edit distance between strings ``bcdea'' and ``abcde'' is two:
insert an 'a' at the front of the first string and delete the 'a' from its end.
The Hamming distance between the same strings is five: every symbol must
be changed.
Edit distance is probably a better way to measure violations of a \reg\
constraint. 
We provide a more natural example in the area of rostering.
Given a string, we call \emph{stretch} a maximal substring of identical values.
We often need to impose restrictions on the length of stretches of work shifts,
and these can be expressed with a \reg\ constraint.
Suppose stretches of $a$'s and $b$'s must each be of length $2$ and
consider the string ``abbaabbaab'': 
its Hamming distance to a string belonging to the corresponding regular
language is $5$ since changing either the first $a$ to a $b$ or $b$
to an $a$ has a domino effect on the following stretches; 
its edit distance is just $2$ since we can insert an $a$ at 
the beginning to make a legal stretch of $a$'s and remove the $b$ at the
end.
In this case, the edit distance reflects the number of illegal
stretches whereas the Hamming distance is proportional to the length of
the string.

\subsection{Cost Evaluation and Cost-Based Filtering}

\begin{figure}
\begin{center}
~
\psfig{figure=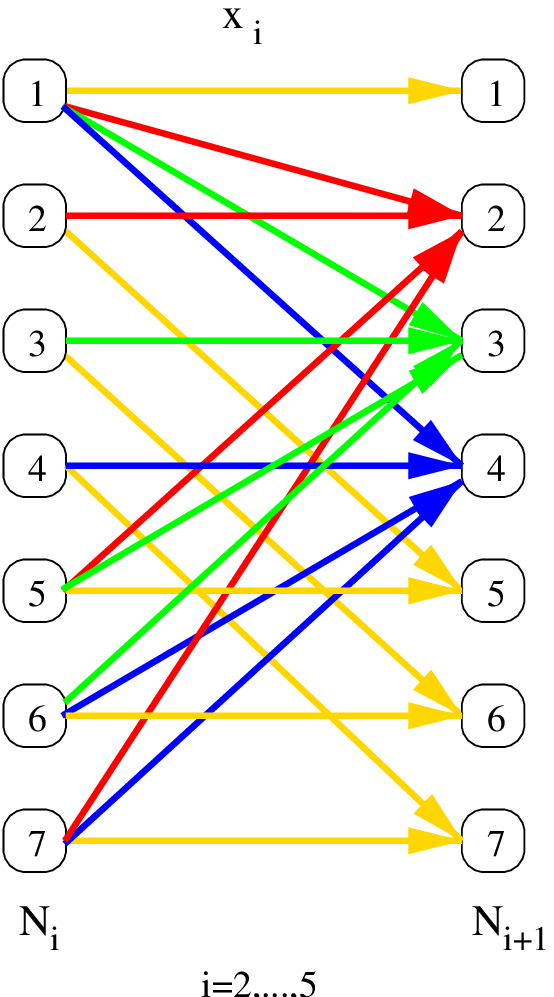,height=5.5cm}
\hspace{4em}
\psfig{figure=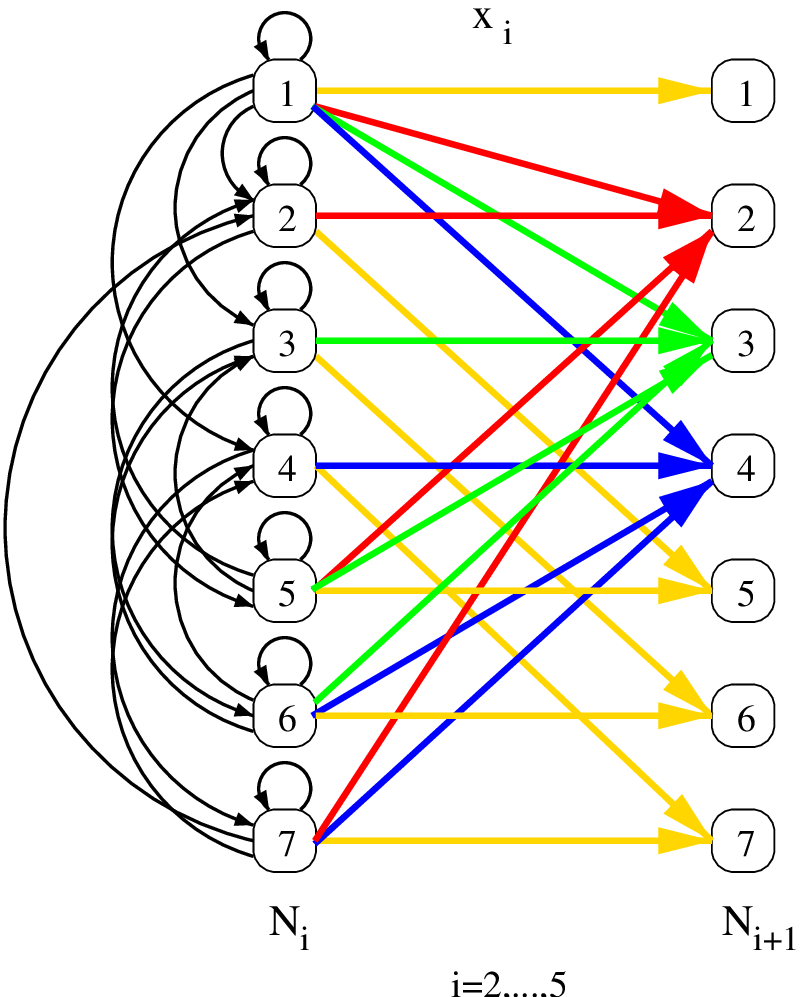,height=5.5cm}
\end{center}
\caption{Shorthand versions of ${\cal G}_{\rm var}$ (left) and ${\cal G}_{\rm edit}$
(right) for the DFA of Figure \ref{digraph}.}
\label{soft-digraph}
\end{figure}

For both cost measures, we proceed by modifying the layered directed
graph ${\cal G}$ built for the ``hard'' version of \reg\ into graph
${\cal G}_{\rm var}$. 
Before, we added an arc from $q^i_k$ to $q^{i+1}_{\ell}$ if $\delta(q_k,v_j) =
q_{\ell}$ for some $v_j \in D_i$;
now we relax it slightly to any $v_j \in \Sigma$.
This only makes a difference if the domains of the variables are not
initially full.
Arcs are never removed in ${\cal G}_{\rm var}$ but their labels are updated instead.
The label of an arc $(q^i_k,q^{i+1}_{\ell})$ is generalized to
the invariant $V_{ik\ell} = \{v_j \in D_i \; | \; \delta(q_k,v_j) =
q_{\ell}\}$; as values are removed from the domain of variable $x_i$,
they are also removed from the corresponding $V_{ik\ell}$'s.
The cost of using an arc $(q^i_k,q^{i+1}_{\ell})$ for variable-value
pair $\langle x_i,v_j \rangle$ will be zero if $v_j$ belongs to
$V_{ik\ell}$ and some positive integer cost otherwise.
This cost represents the penalty for an individual violation.
In the remainder of the section we will consider unit costs but the
framework also makes it possible to use varying costs, e.g. to
distinguish between insertions and substitutions when using the edit
distance.
The graph on the left at Figure \ref{soft-digraph} is a shorthand
version of ${\cal G}_{\rm var}$ for the automaton of Figure \ref{digraph}.
Since all values in $\Sigma$ are considered, the same arcs appear
between consecutive layers.
What changes from one layer to the other are the $V_{ik\ell}$ labels.

Taking into account \emph{substitutions}, common to both Hamming and edit
distances, is immediate from the previous modification.
It is not difficult to see that the introduction of costs transforms a 
supporting path in the domain consistency algorithm for \reg\ into a
zero-cost path in the modified graph.
The cost of a shortest path from $q_0$ in the first layer to a member
of $F$ in the last layer corresponds to the smallest number of
variables forced to take a value outside of their domain.
\begin{theorem}
\label{thm:cost-eval}
A minimum-cost path from $u \in N^1$ to $v \in N^{n+1}$ in 
${\cal G}_{\rm var}$ corresponds to a solution to $\softreg[{\rm var}]$
minimizing the variable-based cost (Hamming distance).
\end{theorem}
Just as the existence of a path through a given arc representing a
variable-value pair constituted a support for that pair in the
filtering algorithm for \reg, the existence of a path whose cost doesn't exceed $\max{D_z}$ constitutes a support for that variable-value pair in a cost-based
filtering algorithm for \softreg.

\begin{theorem}
\label{thm:cost-filt}
$\softreg[{\rm var}]({\mathbf x},M,z)$ is
domain consistent on $\mathbf x$ and bound consistent on $z$ if and only if
\[\min_{q_f \in F}\{{\rm cost}({\rm SP}(q_0^1, q_f^{n+1}))\} \leq
\min{D_z}\] 
and
\[\min_{q_k \in Q,q_f \in F}\{{\rm cost}({\rm SP}(q_0^1, q_k^i)) +
{\rm cost}({\rm SP}(q_{\ell}^{i+1}, q_f^{n+1}))\} \leq \max{D_z}, \quad
\forall x_i \in {\mathbf x},  v_j \in D_i \]
where $\delta(q_k,v_j) = q_{\ell}$ and ${\rm cost}({\rm SP}(u, v))$
denotes the cost of a shortest path from $u$ to $v$ in ${\cal G}_{\rm var}$.
\end{theorem}

Computing shortest paths from the initial state in the first layer to
every other node and from every node to a final state in the last layer can
be done in $O(n \card{\delta})$ time\footnote{$\card{\delta}$ refers
to the number of transitions in the automaton.}
through topological sorts because
of the special structure of the graph.
That computation can also be made incremental in the same way as in
\cite{regular04}.
Recently, that same result was independently obtained in
\cite{beldi-petit:cp04}. 
We however go further by considering edit distance, for which insertions and
deletions are allowed as well.

For \emph{deletions} we need to allow ``wasting'' a value without changing the
current state.
To this effect, we add to ${\cal G}_{\rm var}$ an arc $(q^i_k,q^{i+1}_k)\; \forall \; 1
\leq i \leq n, q_k \in Q$, with $V_{ikk} = \emptyset$, if
it isn't already present in the graph.
To allow \emph{insertions}, inspired by $\epsilon$-transitions in DFAs, we
introduce some special arcs between nodes in the same layer:
if $\exists v \in \Sigma \mbox{ such that }\delta(q_k,v) = q_{\ell}$
then we further add an arc $(q^i_k,q^i_{\ell})\;
\forall \; 1 \leq i \leq n+1$ with fixed positive cost.
Figure \ref{soft-digraph} provides an example of the resulting graph
(on the right).
Unfortunately, those special arcs modify the structure of the graph
since cycles (of strictly positive cost) are introduced.
Consequently shortest paths can no longer be computed through
topological sorts.
An efficient implementation of Dijkstra's algorithm increases the time
complexity to $O(n \card{\delta} + n  \card{Q} \log (n  \card{Q}))$.
Regardless of this increase in computational cost, Theorems
\ref{thm:cost-eval} and \ref{thm:cost-filt} can be generalized to hold
for \softreg[edit] as well.

\section{Aggregating Soft Constraints}
\label{agg}

The preceding sections have introduced filtering algorithms based on
different violation measures for two soft global constraints. If these
filtering techniques are to be effective, especially in the presence
of soft constraints of a different nature, they must be able to
cooperate and communicate. Even though there are many avenues for 
combining soft constraints, the objective almost always remains 
to minimize constraint violations.
We propose here a small extension to the approach of 
\cite{MetaConstraints}, where meta-constraints on the cost variables 
of soft constraints are introduced. We illustrate this approach with 
the newly introduced \softgcc.

\begin{definition}[Soft global cardinality aggregator] Let ${\cal S}$ 
be a set of soft constraints and $z_i \in D_{z_i}$ the variable indicating 
the violation cost of $S_i \in {\cal S}$. The \emph{soft global cardinality
aggregator ({\tt sgca})} is defined as 
$\softgcc[\star](Z,l,u,z_{\rm agg})$ where 
$Z = \{z_1, \dots, z_{ \card{ {\cal S} }} \}$, $l_i,u_i$ is the interval 
defining the allowed number of occurrences of each value in the domain of 
$z_i$ and $z_{\rm agg} \in D_{z_{\rm agg}} \subseteq \nats$ the 
cost variable based on the violation measure $\star$.
\end{definition}

When all constraints are either satisfied or violated 
($Z \in \{0,1\}^{\card{ {\cal S} }}$) the 
Max-CSP approach can be easily obtained by setting $l_1 = 0$, $u_1 =0$,
 $violation(Z) = \sum_{d \in D_Z} overflow(Z,d)$ and reading the number 
of violations in $z_{\rm agg}$.
The {\tt sgca} could also be used as in \cite{MetaConstraints} to enforce 
homogeneity (in a soft manner) or to define other violation measures like 
restricting the number of highly violated constraint. For instance, we could 
wish to impose that no more then a certain number of constraints are highly 
violated, but 
since we cannot guarantee that this is possible the use of {\tt sgca} allows 
to state this wish without risking to create an inconsistent problem. More 
generally, by defining the values of $l$ and $u$ accordingly it is possible 
to limit (or at least attempt to limit) the number violated constraints by 
violation cost. Another approach could be to set all $u$ to 0 and adjust the 
violation function so that higher violation costs are more penalized. The 
use of soft meta-constraints, when possible, is also an alternative to the 
introduction of disjunctive constraints since they need not be satisfied for 
the problem to be consistent.

In the original meta-constraint framework, similar behaviour can be 
established by applying a cost-\gcc\ to $Z$. For instance, we can define
for each pair $(z_i, d)$ ($d \in D_{z_i}$) a cost $d$ which penalizes 
higher violations more. With the \softgcc, this cost function can be 
stated as ${\rm violation}(Z) = \sum_{d \in D_Z} d \cdot {\rm overflow}(Z,d)$.
However, as for this variant of the \softgcc\ we have $l = \vec{0}$, 
the \softgcc\ will be much more efficient than the cost-\gcc, as was 
discussed at the end of Section~\ref{gcc}. In fact, the {\tt sgca} can be
checked for consistency in $O(nm)$ time and made domain consistent in 
$O(m)$ time (where $n = \card{ {\cal S}}$ and $m = \cup_i \card{D_{z_i}}$
whenever $l = \vec{0}$ and ${\rm violation}(Z) = \sum_{d \in D_Z} F(d) \cdot  
{\rm overflow}(Z,d)$ for any cost function $F: D_Z \rightarrow \reals_+$.


\section{Conclusion}
\label{conclusion}

We have presented soft versions of two global constraints: the global
cardinality constraint and the regular constraint. Different violation
measures have been presented and the corresponding filtering
algorithms achieving domain consistency have been introduced. These
new techniques are based on the addition of ``relaxation arcs'' in the
underlying graph and the use of known graph algorithms. We also have
proposed to extend the Meta-Constraint framework for combining
constraint violations by using the soft version of \gcc. 

Since these two constraints are very useful to solve Personnel
Rostering Problems the next step is thus the implementation of these
algorithms in order to model such problems and benchmark these new
constraints. 


\end{document}